%% file: ms.tex
\documentclass[review, sort&compress]{elsarticle}
\usepackage[breaklinks]{hyperref}
\usepackage{times}
\usepackage{latexsym}

\usepackage{geometry}

\usepackage{subfiles}
\usepackage{multirow}
\usepackage{array}
\usepackage{hyphenat}
\usepackage{subcaption}
\usepackage{longtable}
\usepackage{graphicx}
\usepackage{adjustbox}
\usepackage{enumitem}
\usepackage{url}

\usepackage[color=yellow, textsize=small]{todonotes}

\usepackage{bm}
\usepackage{booktabs}
\usepackage{float}
\usepackage[english]{babel}
\usepackage{blindtext}
\usepackage{textcomp}
\usepackage{soul,color}

\usepackage{epstopdf}
\AppendGraphicsExtensions{.tif}

\usepackage{setspace}
\newif\ifsubfile
\subfiletrue

\newif\iftif
\tiffalse

\usepackage{microtype}

\journal{arXiv}

\begin{document}

\subfilefalse

\begin{frontmatter}

\title{Extracting COVID-19 Diagnoses and Symptoms From Clinical Text: \\ A New Annotated Corpus and Neural Event Extraction Framework}

\author[1]{Kevin Lybarger}\corref{cor1}
\ead{lybarger@uw.edu}

\author[2]{Mari Ostendorf}
\author[3]{Matthew Thompson}
\author[1]{Meliha Yetisgen}

\cortext[cor1]{Corresponding author}

\address[1]{Biomedical \& Health Informatics, University of Washington, Box 358047, Seattle, WA 98109, USA}
\address[2]{Department of Electrical \& Computer Engineering, University of Washington, Campus Box 352500 185, Seattle, WA 98195-2500, USA}
\address[3]{Department of Family Medicine, University of Washington, Box 354696, Seattle, WA 98195-2500, USA}

\begin{abstract}
\subfile{sections/abstract}

\end{abstract}


\end{frontmatter}

\begin{abstract}
\subfile{sections/abstract}
\end{abstract}

\section{Introduction}

\subfile{sections/introduction}

\section{Related Work}
\subfile{sections/related_work}

\section{Materials}

\subfile{sections/materials}

\section{Event Extraction}

\subfile{sections/event_extraction}

\section{COVID-19 Prediction Application}

\subfile{sections/covid_prediction}

\section{Conclusions}

\subfile{sections/conclusions}

\section*{Acknowledgments}
This work was funded by the National Library of Medicine Biomedical and Health Informatics Training Program under Award Number 5T15LM007442-19, the Gordon and Betty Moore Foundation, and the National Center For Advancing Translational Sciences of the National Institutes of Health under Award Number UL1 TR002319.  We want to acknowledge Elizabeth Chang, Kylie Kerker, Jolie Shen, and Erica Qiao for their contributions to the gold standard annotations and Nicholas Dobbins for data management and curation. Research and results reported in this publication was partially facilitated by the generous contribution of computational resources from the University of Washington Department of Radiology.

\bibliography{mybib}

\section{Appendix}

\subfile{sections/appendix}

\end{document}

%% file: sections/abstract.tex
Coronavirus disease 2019 (COVID-19) is a global pandemic. Although much has been learned about the novel coronavirus since its emergence, there are many open questions related to tracking its spread, describing symptomology, predicting the severity of infection, and forecasting healthcare utilization. Free-text clinical notes contain critical information for resolving these questions. Data-driven, automatic information extraction models are needed to use this text-encoded information in large-scale studies. This work presents a new clinical corpus, referred to as the COVID-19 Annotated Clinical Text (CACT) Corpus, which comprises 1,472 notes with detailed annotations characterizing COVID-19 diagnoses, testing, and clinical presentation. We introduce a span-based event extraction model that jointly extracts all annotated phenomena, achieving high performance in identifying COVID-19 and symptom events with associated assertion values (0.83-0.97 F1 for events and 0.73-0.79 F1 for assertions). Our span-based event extraction model outperforms an extractor built on MetaMapLite for the identification of symptoms with assertion values. In a secondary use application, we predicted COVID-19 test results using structured patient data (e.g. vital signs and laboratory results) and automatically extracted symptom information, to explore the clinical presentation of COVID-19. Automatically extracted symptoms improve COVID-19 prediction performance, beyond structured data alone.

%% file: sections/introduction.tex
As of December 20, 2020, there were over 75 million confirmed COVID-19 cases globally, resulting in 1.6 million related deaths \citep{WHO_situation}. Surveillance efforts to track the spread of COVID-19 and estimate the true number of infections remains a challenge for policy makers, healthcare workers, and researchers, even as testing availability increases. Symptom information provides useful indicators for tracking potential COVID-19 infections and disease clusters \citep{rossman2020framework}. Certain symptoms and underlying comorbidities have directed COVID-19 testing. However, the clinical presentation of COVID-19 varies significantly in severity and symptom profiles \citep{wu2020characteristics}. 

The most prevalent COVID-19 symptoms reported to date are fever, cough, fatigue, and dyspnea \citep{yang2020prevalence}, but emerging reports identify additional symptoms, including diarrhea and neurological symptoms, such as changes in taste or smell \citep{vetter2020clinical, qian2020covid, wei2020presymptomatic}. Certain initial symptoms may be associated with higher risk of complications; in one study, dyspnea was associated with a two-fold increased risk of acute respiratory distress syndrome \citep{wu2020risk}. However, correlations between symptoms, positive tests, and rapid clinical deterioration are not well understood in ambulatory care and emergency department settings. 

Routinely collected information in the Electronic Health Record (EHR) can provide crucial COVID-19 testing, diagnosis, and symptom data needed to address these knowledge gaps. Laboratory results, vital signs, and other structured data results can easily be queried and analyzed at scale; however, more detailed and nuanced descriptions of COVID-19 diagnoses, exposure history, symptoms, and clinical decision-making are typically only documented in the clinical narrative. To leverage this textual information in large-scale studies, the salient COVID-19 and symptom information must be automatically extracted.

This work presents a new corpus of clinical text annotated for COVID-19, referred to as the COVID-19 Annotated Clinical Text (CACT) Corpus. CACT consists of 1,472 notes from the University of Washington (UW) clinical repository with detailed event-based annotations for COVID-19 diagnosis, testing, and symptoms. The event-based annotations characterize these phenomena across multiple dimensions, including assertion, severity, change, and other attributes needed to comprehensively represent these clinical phenomena in secondary use applications. This work is part of a larger effort to use routinely collected data describing the clinical presentation of acute and chronic diseases, with two major aims; 1) to describe the presence, character, and changes in symptoms associated with clinical conditions, where delays or misdiagnoses occur in clinical practice and impact patient outcomes (e.g. infectious diseases, cancer) \cite{newman2019serious}, and 2) to provide a more efficient and cost-effective mechanism to validate clinical prediction rules previously derived from large prospective cohort studies \cite{collins2015transparent}. To the best of our knowledge, CACT is the first clinical data set with COVID-19 annotations, and it includes 29.9K distinct events. We present the first information extraction results on CACT using an end-to-end neural event extraction model, establishing a strong baseline for identifying COVID-19 and symptom events. We explore the prediction of COVID-19 test results (positive or negative) using structured EHR data and automatically extracted symptoms and find that the automatically extracted symptoms improve prediction performance.

\ifsubfile
\bibliography{mybib}
\fi

%% file: sections/related_work.tex
\label{lit_review}

\subsection{Annotated Corpora}
Given the recent onset of COVID-19, there are limited COVID-19 corpora  for natural language processing (NLP) experimentation. Corpora of scientific papers related to COVID-19 are available \citep{Wang2020CORD19TC, WHO_corpora}, and automatic labels for biomedical entity types are available for some of these research papers \citep{wang2020comprehensive}. However, we are unaware of corpora of clinical text with supervised COVID-19 annotations.

Multiple clinical corpora are annotated for symptoms. As examples, \citet{south2009developing} annotated symptoms and other medical concepts with  negation (present/not present), temporality, and other attributes. \citet{koeling2011annotating} annotated a pre-defined set of symptoms related to ovarian cancer. For the i2b2/VA challenge, \citet{uzuner20112010} annotated medical concepts, including symptoms, with assertion values and relations. While some of these corpora may include symptom annotations relevant to COVID-19 (e.g. ``cough'' or ``fever''), the distribution and characterization of symptoms in these corpora may not be consistent with COVID-19 presentation. To fill the gap in clinical COVID-19 annotations and detailed symptom annotation, we introduce CACT to provide a relatively large corpus with COVID-19 diagnosis, testing, and symptom annotations.

\subsection{Medical Concept and Symptom Extraction}
The most commonly used United Medical Language System (UMLS) concept extraction systems are the clinical Text Analysis and Knowledge Extraction System (cTAKES) \cite{savova2010mayo} and MetaMap \cite{aronson2001effective}. The National Library of Medicine (NLM) created a lightweight Java implementation of MetaMap, \textit{MetaMapLite}, which demonstrated real-time speed and extraction performance comparable to or exceeding the performance of MetaMap, cTAKES, and DNorm \cite{demner2017metamap}. In previous work, we built on MetaMapLite, incorporating assertion value predictions (e.g. present versus absent) using classifiers trained on the 2010 i2b2 challenge dataset to create the extraction pipeline referred to here as \textit{MetaMapLite++} \cite{yetisgen2016new}. MetaMapLite++ assigns each extracted UMLS Metathesaurus concept an assertion value with an Support Vector Machine (SVM)-based assertion classifier that utilizes syntactic and semantic knowledge. The SVM assertion classifier achieved state-of-the-art assertion performance (Micro-F1 94.23) on the i2b2 2010 assertion dataset \cite{bejan2013assertion}. Here, we use MetaMapLite++ as a baseline for evaluating extraction performance for a subset of our annotated phenomena, specifically symptoms with assertion values, using the UMLS ``Sign or Symptom'' semantic type. The Mayo Clinic updated its rule-based medical tagging system, MedTagger \citep{wen2019desiderata}, to include a COVID-19 specific module that extracts 18 phenomena related to COVID-19, including 11 common COVID-19 symptoms with assertion values \cite{ohnlp}. We do not use the COVID-19 MedTagger variant as a baseline, because our symptom annotation and extraction is not limited to known COVID-19 symptoms.

\subsection{Relation and Event Extraction}
There is a significant body of information extraction (IE) work related to coreference resolution, relation extraction, and event extraction tasks. In these tasks, spans of interest are identified, and linkages between spans are predicted. Many contemporary IE systems use end-to-end multi-layer neural models that encode an input word sequence using recurrent or transformer layers, classify spans (entities, arguments, etc.), and predict the relationship between spans (coreference, relation, role, etc.) \citep{ZHENG201759, orr2018event, Shi_2019_extracting, Pang2019deep, chen_2019_extracting, christopoulou2020adverse}. Of most relevance to our work is a series of developments starting with \citet{lee-etal-2017-end}, which introduces a span-based coreference resolution model that enumerates all spans in a word sequence, predicts entities using a feed-forward neural network (FFNN) operating on span representations, and resolves coreferences using a FFNN operating on entity span-pairs. \citet{luan-2018-multi} adapts this framework to entity and relation extraction, with a specific focus on scientific literature. \citet{luan2019general} extends the method to take advantage both of co-reference and relation links in a graph-based approach to jointly predict entity spans, co-reference, and relations. By updating span representations in multi-sentence co-reference chains, the graph-based approach achieved state-of-the-art on several IE tasks representing a range of different genres. \citet{wadden2019entity} expands on \citet{luan2019general}'s approach, adapting it to event extraction tasks. We build on \citet{luan-2018-multi} and \citet{wadden2019entity}'s work, augmenting the modeling framework to fit the CACT annotation scheme. In CACT, event arguments are generally close to the associated trigger, and inter-sentence events linked by co-reference are infrequent, so the graph-based extension, which adds complexity, is unlikely to benefit our extraction task.

Many recent NLP systems use pre-trained language models (LMs), such as ELMo, BERT, and XLNet, that leverage unannotated text \cite{peters2018deep, devlin2019bert, yang2019xlnet}. A variety of strategies for incorporating the LM output are used in IE systems, including using the contextualized word embedding sequence: as the input to a Conditional Random Field entity extraction layer \citep{HuangW2019BMSf}, as the basis for building span representations \citep{luan2019general,wadden2019entity}, or by adding an entity-aware attention mechanism and pooled output states to a fully transformer-based model \citep{wang-etal-2019-extracting}. There are many domain-specific LM variants. Here, we use \citet{alsentzer-etal-2019-publicly}'s \textit{Bio+Clinical BERT}, which is trained on PubMed papers and MIMIC-III \citep{johnson2016mimic} clinical notes.

\subsection{COVID-19 Outcome Prediction}

There are many pre-print and published works exploring the prediction of COVID-19 outcomes, including COVID-19 infection, hospitalization, acute respiratory distress syndrome, need for intensive care unit (ICU), need for a ventilator, and mortality \citep{tian2020predictors, figliozzi2020predictors, jain2020predictive, dong2020novel, xu2020risk, Izquierdo2020, bertsimas2020predictions, Wynantsm1328, SIORDIA2020104357,   zhang2020risk, Brinati2020, Mei2020, wollenstein2020personalized}. These COVID-19 outcomes are typically predicted using existing structured data within the EHR, including demographics, diagnosis codes, vitals, and lab results, although \citet{Izquierdo2020} incorporates automatically extracted information from the existing EHRead tool. Our literature review identified 24 laboratory, vital sign, and demographic fields that are predictive of COVID-19 (see Table \ref{lit_features} in the Appendix details). While there are some frequently cited fields,  there does not appear to be a consensus across the literature regarding the most prominent predictors of COVID-19 infection. These 24 predictive fields informed the development of our COVID-19 prediction work in Section \ref{covid_prediction}. Prediction architectures includes logistic regression, SVM, decision trees, random forest, K-nearest neighbors, Na\"ive Bayes, and multilayer perceptron \citep{Izquierdo2020, bertsimas2020predictions, Brinati2020, Mei2020, wollenstein2020personalized}.

\ifsubfile
\bibliography{mybib}
\fi

%% file: sections/materials.tex
\subsection{Data}
This work used inpatient and outpatient clinical notes from the UW clinical repository. COVID-19-related notes were identified by searching for variations of the terms \textit{coronavirus}, \textit{covid}, \textit{sars-cov}, and \textit{sars-2} in notes authored between February 20-March 31, 2020, resulting in a pool of 92K notes. Samples were randomly selected for annotation from a subset of 53K notes that include at least five sentences and correspond to the note types: telephone encounters, outpatient progress, emergency department, inpatient nursing, intensive care unit, and general inpatient medicine. Multiple note types were used to improve extraction model generalizability.

Early in the outbreak, the UW EHR did not include COVID-19 specific structured data; however, structured fields indicating COVID-19 test types and results were added as testing expanded. We used these structured fields to assign a \textit{COVID-19 Test} label describing COVID-19 polymerase chain reaction (PCR) testing to each note based on patient test status within the UW system (no data external to UW was used):
\begin{itemize}[nolistsep]
    \item \textit{none}: patient testing information is not available
    \item \textit{positive}: patient will have at least one future positive test
    \item \textit{negative}: patient will only have future negative tests
\end{itemize}
\noindent
More nuanced descriptions of COVID-19 testing (e.g. conditional or unordered tests) or diagnoses (e.g. possible infection or exposure) are not available as structured data. For the 53K note subset, the \textit{COVID-19 Test} label distribution is 90.8\% \textit{none}, 7.9\% \textit{negative}, and 1.3\% \textit{positive}.\footnote{The COVID-19 test positivity rate cannot be inferred from these label distributions, as there can be multiple test results associated with each note-level label.}

Given the sparsity of \textit{positive} and \textit{negative} notes, CACT is intentionally biased to increase the prevalence of these labels. To ensure adequate \textit{positive} training samples, the CACT training partition includes 46\% \textit{none}, 5\% \textit{negative}, and 49\% \textit{positive} notes. Ideally, the test set would be representative of the true distribution; however, the expected number of \textit{positive} labels with random selection is insufficient to evaluate extraction performance. Consequently, the CACT test partition was biased to include 50\% \textit{none}, 46\% \textit{negative}, and 4\% \textit{positive} notes. Notes were randomly selected in equal proportions from the six note types. CACT includes 1,472 annotated notes, including 1,028 train and 444 test notes.

\subsection{Annotation Scheme}
\begin{table*}[htb!]
    \small
    \centering

\input{tables/annotated_phenomena.tex}
    \caption{Annotation guideline summary. \textsuperscript{*} indicates the argument is required. \textsuperscript{$\dagger$} indicates at least one of the arguments, \textit{Test Status} or \textit{Assertion}, is required}
    \label{annotated_phenomena}
\end{table*}

We created detailed annotation guidelines for COVID-19 and symptoms, using the event-based annotation scheme in Table \ref{annotated_phenomena}. Each event includes a trigger that identifies and anchors the event and arguments that characterize the event. The annotation scheme includes two types of arguments: \textit{labeled arguments} and \textit{span-only arguments}. \textit{Labeled arguments} (e.g. \textit{Assertion}) include an argument span, type, and subtype (e.g. \textit{present}). The subtype label normalizes the span information to a fixed set of classes and allows the extracted information to be directly used in secondary use applications. \textit{Span-only arguments} (e.g. \textit{Characteristics}) include an argument span and type but do not include a subtype label, because the argument information is not easily mapped to a fixed set of classes. 

\begin{figure}[ht]
\begin{adjustbox}{varwidth=\textwidth,fbox,center}
\begin{subfigure}{3.0in}
    \iftif
        \includegraphics[scale=0.20]{figures/brat_example1.tif}  
    \else 
        \includegraphics[scale=0.20]{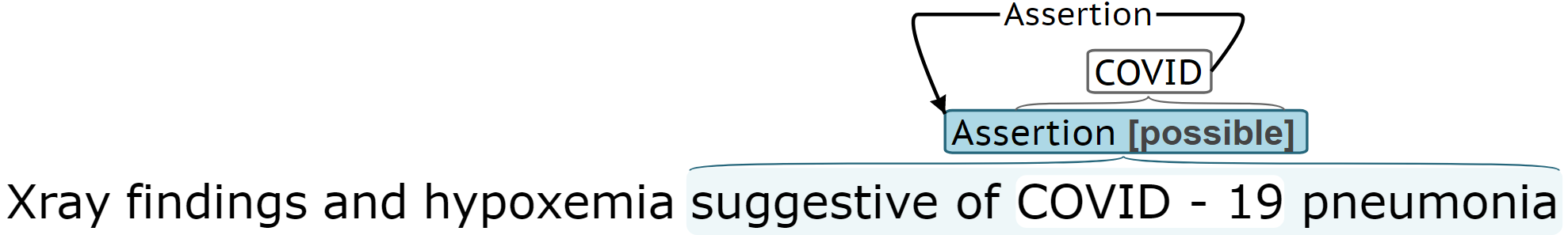}  
    \fi
\end{subfigure}
\par\medskip
\begin{subfigure}{3.0in}
    \iftif
        \includegraphics[scale=0.20]{figures/brat_example2.tif}      
    \else
        \includegraphics[scale=0.20]{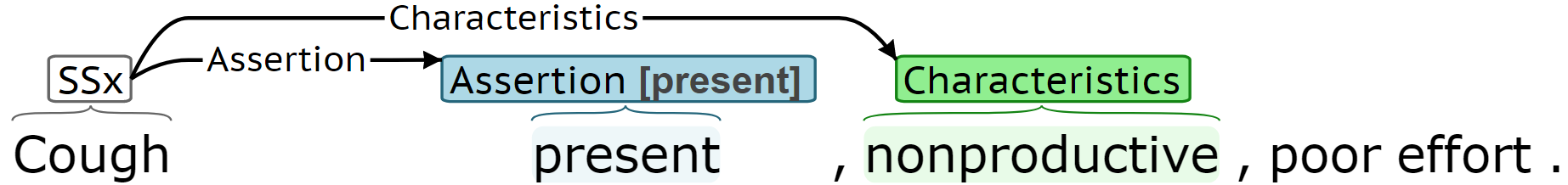}      
    \fi
\end{subfigure}
\end{adjustbox}
\caption{BRAT annotation examples for \textit{COVID} and \textit{Symptom} (\textit{SSx}) event types}
\label{annotation_examples}
\end{figure}

For \textit{COVID} events, the trigger is generally an explicit COVID-19 reference, like ``COVID-19" or ``coronavirus." \textit{Test Status} characterizes implicit and explicit references to COVID-19 testing, and \textit{Assertion} captures diagnoses and hypothetical references to COVID-19.  \textit{Symptom} events capture subjective, often patient reported, indications of disorders and diseases (e.g ``cough''). For \textit{Symptom} events, the trigger identifies the specific symptom, for example ``wheezing" or ``fever," which is characterized through \textit{Assertion}, \textit{Change}, \textit{Severity}, \textit{Anatomy}, \textit{Characteristics}, \textit{Duration}, and \textit{Frequency} arguments. Symptoms were annotated for all conditions/diseases, not just COVID-19. Notes were annotated using the BRAT annotation tool \citep{stenetorp2012brat}. Figure \ref{annotation_examples} presents BRAT annotation examples.

Most prior medical problem extraction work, including symptom extraction, focuses on identifying the specific problem, normalizing the extracted phenomenon, and predicting an assertion value (e.g. present versus absent). This approach omits many of the symptom details that clinicians are taught to document and that form the core of many clinical notes. This symptom detail describes change (e.g. improvement, worsening, lack of change), severity (e.g. intensity and impact on daily activities), particular characteristics  (e.g. productive, dry, or barking for cough), and location. We hypothesize that this symptom granularity is needed for many clinical conditions to improve timely diagnosis and validate diagnosis prediction rules.

\subsection{Annotation Scoring and Evaluation}
\begin{figure}[ht]
\begin{adjustbox}{varwidth=\textwidth,fbox,center, padding=0ex 0ex 0ex 0ex, margin=0ex 0ex 0ex 0ex}
\begin{subfigure}{3.0in}
  \centering
    \iftif
        \includegraphics[scale=0.20]{figures/slot_filling.tif}
    \else
        \includegraphics[scale=0.20]{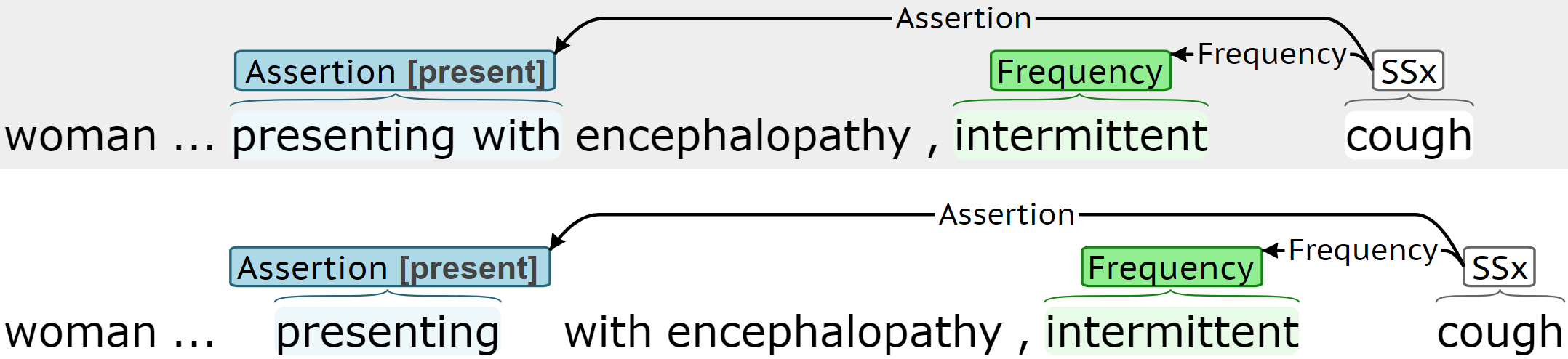}     
    \fi
\end{subfigure}
\par\medskip
\begin{subfigure}{3.0in}
\centering
\large
$\bm{\Downarrow}$ \\
\footnotesize
SSx(trigger=``cough'',
\textit{Assertion}=\textit{present},
\textit{Frequency}=``intermittent'')

\end{subfigure}
\end{adjustbox}
\caption{Annotation examples describing event extraction as a slot filling task}
\label{annotation_comparison}
\end{figure}

Annotation and extraction is scored as a slot filling task, focusing on information most relevant to secondary use applications. Figure \ref{annotation_comparison} presents the same sentence annotated by two annotators, along with the populated slots for the \textit{Symptom} event. Both annotations include the same trigger and \textit{Frequency} spans (``cough'' and ``intermittent'', respectively). The \textit{Assertion} spans differ (``presenting with'' vs. ``presenting''), but the assigned subtypes (\textit{present}) are the same, so the annotations are equivalent for purposes of populating a database. Annotator agreement and extraction performance are assessed using scoring criteria that reflects this slot filling interpretation of the labeling task. 

The \textit{Symptom} trigger span identifies the specific symptom. For \textit{COVID}, the trigger anchors the event, although the span text is not salient to downstream applications. For labeled arguments, the subtype label captures the most salient argument information, and the identified span is less informative. For span-only arguments, the spans are not easily mapped to a fixed label set, so the selected span contains the salient information. Performance is evaluated using precision (P), recall (R), and F1.

\textbf{Trigger:} Triggers, $T_i$, are represented by a pair (event type, $e_i$; token indices, $x_i$). Trigger equivalence is defined as
\[
T_i \equiv T_j \mbox{ if } (e_i \equiv e_j) \land (x_i \equiv x_j).
\]

\textbf{Arguments:} Events are aligned based on trigger equivalence. The arguments of events with equivalent triggers are compared using different criteria for \textit{labeled arguments} and \textit{span-only arguments}. Labeled arguments, $L_i$, are represented as a triple (argument type, $a_i$; token indices, $x_i$; subtype, $l_i$). For labeled arguments, the argument type, $a$, and subtype, $l$, capture the salient information and equivalence is defined as 
\[
L_i \equiv L_j \mbox{ if } (T_i \equiv T_j) \land (a_i \equiv a_j) \land (l_i \equiv l_j).
\]
Span-only arguments, $S_i$, are represented as a pair (argument type, $a_i$; token indices, $x_i$). Span-only arguments with equivalent triggers and argument types, $(T_i \equiv T_j) \land (a_i \equiv a_j)$, are compared at the token-level (rather than the span-level) to allow partial matches. Partial match scoring is used as partial matches can still contain useful information. 

\subsection{Annotation Statistics}
\begin{figure}[ht]
    \centering
    \iftif
        \frame{\includegraphics[scale=1.0]{figures/covid_histogram.tif}}
    \else
        \frame{\includegraphics[scale=1.0]{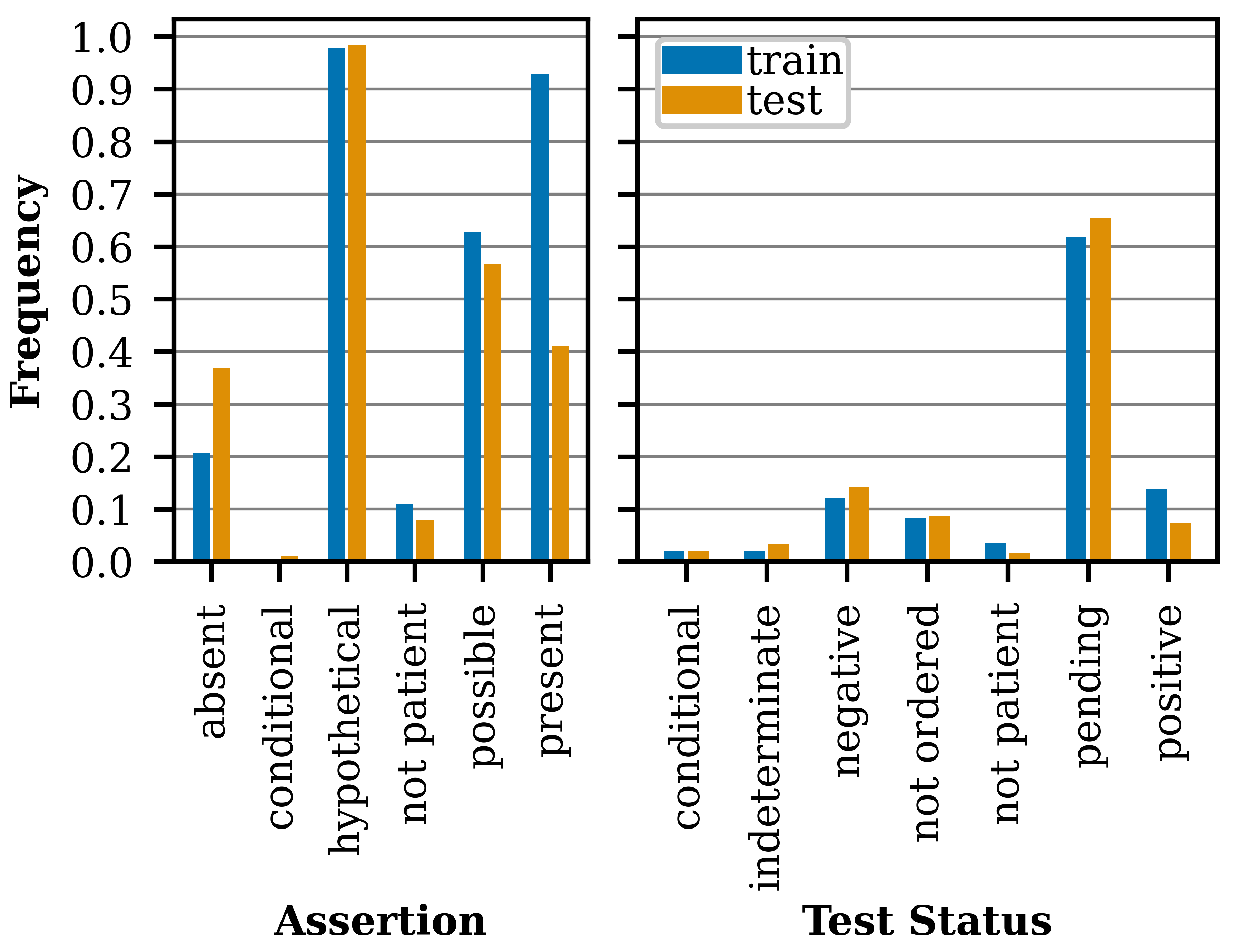}}   
    \fi
    
    \caption{COVID annotation summary}
    \label{covid_stats}
\end{figure}

CACT includes 1,472 notes with a 70\%/30\% train/test split and 29.9K annotated events (5.4K \textit{COVID} and 24.4K \textit{Symptom}). Figure \ref{covid_stats} contains a summary of the \textit{COVID} annotation statistics for the train/test subsets. By design, the training and test sets include high rates of COVID-19 infection (\textit{present} subtype for \textit{Assertion} and \textit{positive} subtype for \textit{Test Status}), with higher rates in the training set. CACT includes high rates of \textit{Assertion} \textit{hypothetical} and \textit{possible} subtypes. The \textit{hypothetical} subtype applies to sentences like, ``She is mildly concerned about the coronavirus'' and ``She cancelled nexplanon replacement due to COVID-19.'' The \textit{possible} subtype applies to sentences like, ``risk of Covid exposure'' and ``Concern for respiratory illness (including COVID-19 and influenza).'' \textit{Test Status} \textit{pending} is also frequent.

There is some variability in the endpoints of the annotated \textit{COVID} trigger spans (e.g. ``COVID'' vs. ``COVID test''); however 98\% of the \textit{COVID} trigger spans in the training set start with the tokens ``COVID,'' ``COVID19,'' or ``coronavirus.'' Since the \textit{COVID} trigger span is only used to anchor and disambiguate events, the \textit{COVID} trigger spans were truncated to the first token of the annotated span in all experimentation and results.

\begin{figure}[ht]
    \centering
    \iftif
        \frame{\includegraphics[scale=1.0]{figures/ssx_histogram.tif}}
    \else
        \frame{\includegraphics[scale=1.0]{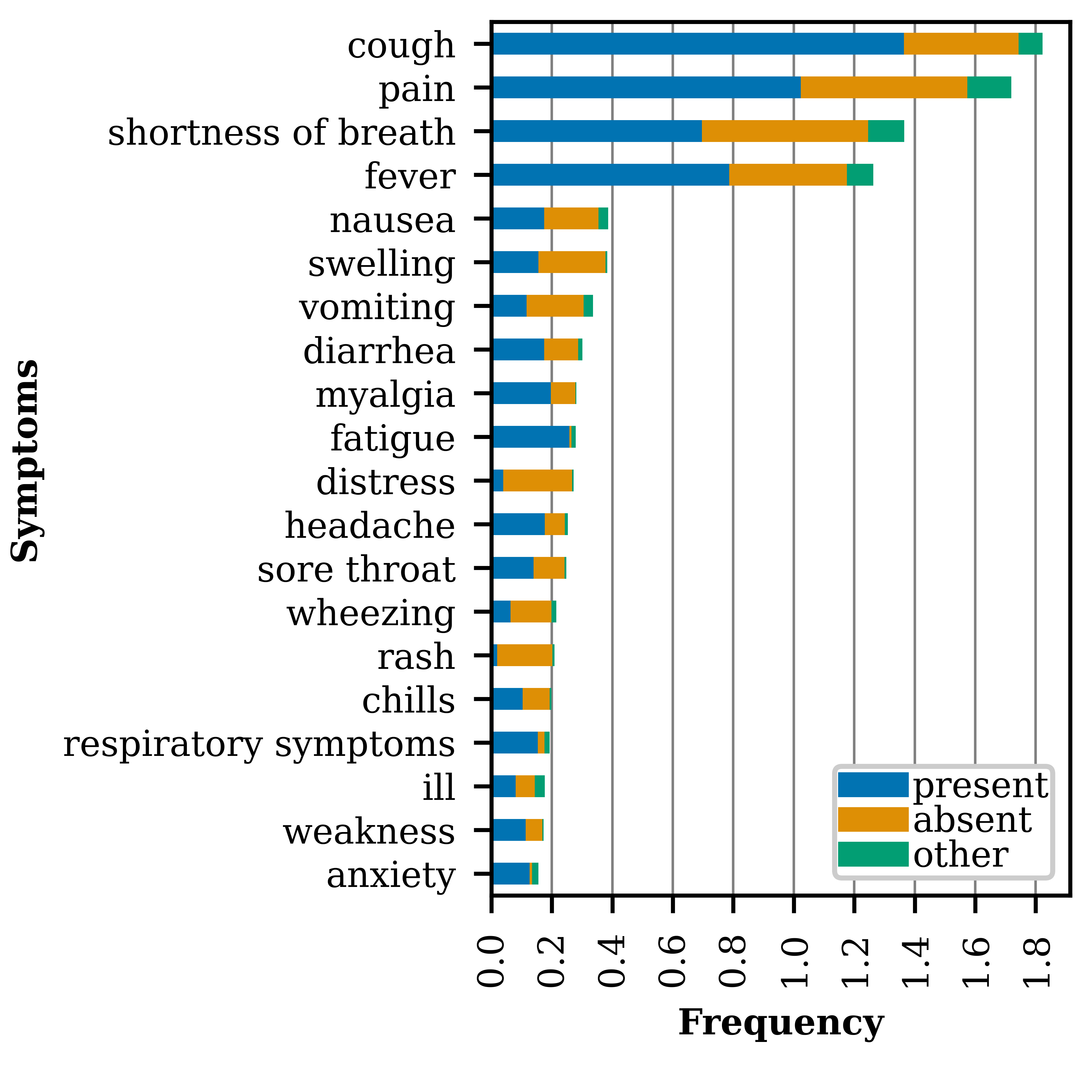}}   
    \fi    
    \caption{Most frequent symptoms in the training set broken down by \textit{Assertion} subtype}
    \label{ssx_histogram}
\end{figure}

The training set includes 1,756 distinct uncased \textit{Symptom} trigger spans, 1,425 of which occur fewer than five times. Figure \ref{ssx_histogram} presents the frequency of the 20 most common \textit{Symptom} trigger spans in the training set by \textit{Assertion} subtypes \textit{present}, \textit{absent}, and other (\textit{possible}, \textit{conditional}, \textit{hypothetical}, or \textit{not patient}). The extracted symptoms in Figure \ref{ssx_histogram} were manually normalized to aggregate different extracted spans with similar meanings (e.g. ``sob'' and ``short of breath'' $\rightarrow$ ``shortness of breath''; ``febrile'' and ``fevers''  $\rightarrow$ ``fever''). Table \ref{symptom_normalization} in the Appendix presents the the symptom normalization mapping, provided by a medical doctor. These 20 symptoms account for 62\% of the training set \textit{Symptom} events. There is ambiguity in delineating between some symptoms and other clinical phenomena (e.g. exam findings and medical problems), which introduces some annotation noise.

Given the long tail of the symptom distribution and our desire to understand the more prominent COVID-19 symptoms, we focused annotator agreement assessment and extraction model training/evaluation on the symptoms that occurred at least 10 times in the training set, resulting in 185 distinct, unnormalized symptoms that cover 82\% of the training set \textit{Symptom} events. The set of 185 symptoms was determined only using the training set, to allow unbiased experimentation on the test set. The subsequent annotator agreement and information extraction experimentation only incorporate these 185 most frequent symptoms.

\subsection{Annotator Agreement}
\begin{figure}[ht]
    \small
    \centering
    \iftif
        \frame{\includegraphics[scale=1.0]{figures/agreement.tif}}
    \else
        \frame{\includegraphics[scale=1.0]{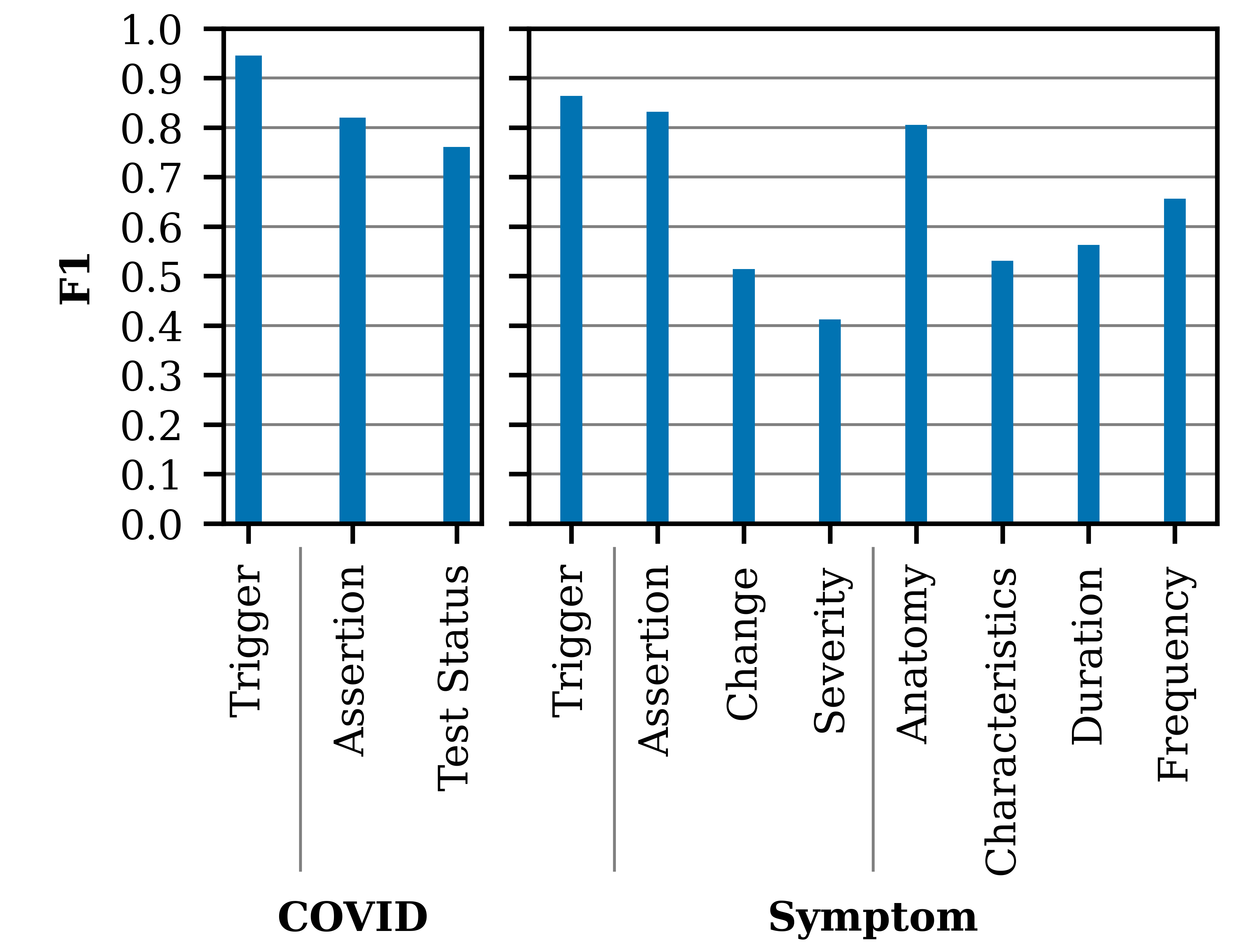}}   
    \fi    
    
    \caption{Annotator agreement}
    \label{agreement}
\end{figure}

 All annotation was performed by four UW medical students in their fourth year. After the first round of annotation, annotator disagreements were carefully reviewed, the annotation guidelines were updated, and annotators received additional training. Additionally, potential \textit{COVID} triggers were pre-annotated using pattern matching (``COVID,'' ``COVID-19,'' ``coronavirus,'' etc.), to improve the recall of \textit{COVID} annotations. Pre-annotated \textit{COVID} triggers were modified as needed by the annotators, including removing, shifting, and adding trigger spans. Figure \ref{agreement} presents the annotator agreement for the second round of annotation, which included 96 doubly annotated notes. For labeled arguments, F1 scores are micro-average across subtypes. 

\ifsubfile
\bibliography{mybib}
\fi

%% file: tables/annotated_phenomena.tex
\begin{tabular}{m{1.8cm} m{2.35cm} m{5.5cm} m{3.8cm}}
\toprule
\textbf{Event type, $e$}                                 & \textbf{Argument type, $a$}                 & \textbf{Argument subtypes, $L_l$}                                                                    & \textbf{Span examples}              \\ \toprule
\multirow{3}{*}{COVID}                        & Trigger\textsuperscript{*}        & --                                                                                      & ``COVID," ``COVID-19"                                     \\ \cmidrule{2-4} 
                                                    & Test Status\textsuperscript{$\dagger$}    & \{positive, negative, pending, conditional, \mbox{not ordered}, \mbox{not patient}, indeterminate \}     & ``tested positive"                   \\ \cmidrule{2-4} 
                                                    & Assertion\textsuperscript{$\dagger$}      & \{present, absent, possible, hypothetical, \mbox{not patient}\}                                  & ``positive," ``low suspicion"                             \\ \midrule
\multirow{7}{*}{Symptom}                 & Trigger\textsuperscript{*}          & --                                                                                    & ``cough," ``shortness of breath"                          \\ \cmidrule{2-4} 
                                                    & Assertion\textsuperscript{*}      & \{\nohyphens{present, absent, possible, conditional, hypothetical, not patient}\}         & ``admits," ``denies"                                      \\ \cmidrule{2-4} 
                                                    & Change                            & \{no change, worsened, improved, resolved\}                                               & ``improved," ``continues"                                 \\ \cmidrule{2-4}                                                     
                                                    & Severity                          & \{mild, moderate, severe\}                                                                & ``mild," ``required ventilation"                          \\ \cmidrule{2-4} 
                                                    & Anatomy                           & --                                                                                    & ``chest wall," ``lower back"                              \\ \cmidrule{2-4} 
                                                    & Characteristics                   & --                                                                                    & ``wet productive," ``diffuse"                             \\ \cmidrule{2-4} 
                                                    & Duration                          & --                                                                                    & ``for two days," ``1 week"                        \\ \cmidrule{2-4}                                                                                                                                                                                     
                                                    & Frequency                         & --                                                                                    & ``occasional," ``chronic"                                 \\ \bottomrule                                                                                                                                                    
\end{tabular}

%% file: sections/event_extraction.tex
\subsection{Methods}
\label{methods_section}

Event extraction tasks, like ACE05 \citep{ace2005}, typically require prediction of the following event phenomena:
\begin{itemize}[nolistsep]
    \item trigger span identification
    \item trigger type (event type) classification
    \item argument span identification
    \item argument type/role classification
\end{itemize}
The CACT annotation scheme differs from this configuration in that labeled arguments require the argument type (e.g. \textit{Assertion}) and the subtype (e.g. \textit{present}, \textit{absent}, etc.) to be predicted. Resolving the argument subtypes require a classifier with additional predictive capacity. 

\begin{figure}[ht!]
    \centering
    \iftif
        \frame{\includegraphics[width=2.91in]{figures/extraction_model.tif}}
    \else
        \frame{\includegraphics[width=2.91in]{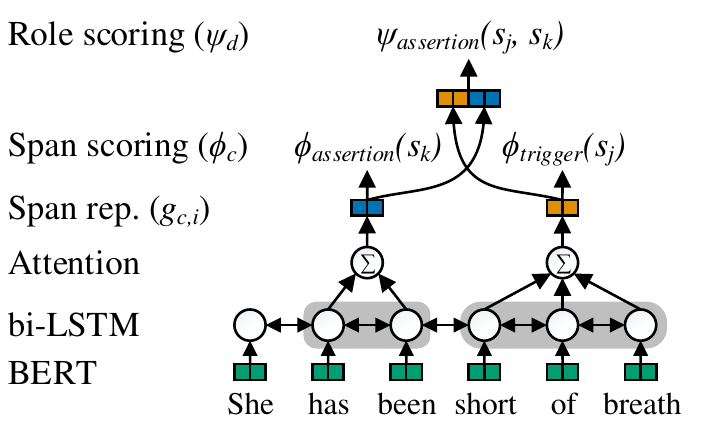}}   
    \fi
    \caption{Span-based Event Extractor}
    \label{extraction_model}
\end{figure}

We implement a span-based, end-to-end, multi-layer event extraction model that jointly predicts all event phenomena, including the trigger span, event type, and argument spans, types, and subtypes. Figure \ref{extraction_model} presents our Span-based Event Extractor framework, which differs from prior related work in that multiple span classifiers are used to accommodate the argument subtypes. 

Each input sentence consists of tokens, $X=\{x_1, x_2, . .. x_n\}$, where $n$ is the number of tokens. For each sentence, the set of all possible spans, $S=\{s_1, s_2, . .. s_m\}$, is enumerated, where $m$ is the number of spans with token length less than or equal to $M$ tokens. The model generates trigger and argument predictions for each span in $S$ and predicts the pairing between arguments and triggers to create events from individual span predictions. 

\textbf{Input encoding:} Input sentences are mapped to contextualized word embeddings using \textit{Bio+Clinical BERT} \citep{alsentzer-etal-2019-publicly}. To limit computational cost, the contextualized word embeddings feed into a bi-LSTM without fine tuning BERT (no backpropagation to BERT). The bi-LSTM has hidden size $v_h$. The forward and backward states, $\bm{h}_{t,f}$ and $\bm{h}_{t,b}$, are concatenated to form the $1 \times 2v_h$ dimensional vector  $\bm{h}_t=[\bm{h}_{t,f}, \bm{h}_{t,b}]$, where $t$ is the token position.

\textbf{Span representation:} Each span is represented as the attention weighted sum of the bi-LSTM hidden states. Separate attention mechanisms, $c$, are implemented for trigger and each labeled argument, and a single attention mechanism is implemented for all span-only arguments, $c \in \{1,2\ldots6\}$ (1 for trigger, 4 for labeled arguments, and 1 for span-only arguments). The attention score for span representation $c$ at token position $t$ is calculated as
\begin{equation}
\alpha_{c,t} =\bm{w}_{\alpha,c} \bm{h}_t^T
\end{equation}
where $\bm{w}_{\alpha,c}$ is a learned $1 \times 2v_h$ vector. For span representation $c$, span $i$, and token position $t$, the attention weights are calculated by normalizing the attention scores as
\begin{equation}
a_{c,i,t} =\frac{\mbox{exp}(\alpha_{c,t})}{\sum\limits_{k = start(s_i)}^{end(s_i)} \mbox{exp}(\alpha_{c,k})},
\end{equation}
where $start(s_i)$ and $end(s_i)$ denote the start and end token indices of span $s_i$. Span representation $c$ for span $i$ is calculated as the attention-weighted sum of the bi-LSTM hidden state as
\begin{equation}
\bm{g}_{c,i} = \sum\limits_{t = start(s_i)}^{end(s_i)} a_{c,i,t}\bm{h}_t.
\end{equation}

\textbf{Span prediction:} Similar to the span representations, separate span classifiers, $c$, are implemented for trigger and each labeled argument, and a single classifier predicts all span-only arguments, $c \in \{1,2\ldots6\}$ (1 for trigger, 4 for labeled arguments, and 1 for span-only arguments). Label scores for classifier $c$ and span $i$ are calculated as
\begin{equation}
\phi_{c}(s_i) = \bm{w}_{s,c}\mbox{FFNN}_{s,c}(\bm{g}_{c,i}),
\end{equation}
where $\phi_{c}(s_i)$ yields a vector of label scores of size $|L_{c}|$, $\mbox{FFNN}_{s,c}$ is a non-linear projection from size $2v_h$ to $v_s$, and $\bm{w}_{s,c}$ has size $|L_{c}| \times v_s$. The trigger prediction label set is $L_{trigger}=\{null, \textit{COVID}, \textit{Symptom}\}$. Separate classifiers are used for each labeled argument (\textit{Assertion}, \textit{Change}, \textit{Severity}, and \textit{Test Status}) with label set, $L_{c}=\{null \cup L_l\}$, where $L_l$ is defined in Table \ref{annotated_phenomena}.\footnote{The assertion classifier uses the larger label set associateed with \textit{Symptom}.} For example, $L_{Severity}=\{null, mild, moderate, severe\}$. A single classifier predicts all span-only arguments with label set, $L_{span-only}= \allowbreak \{null, \allowbreak Anatomy, \allowbreak Characteristics, \allowbreak Duration, \allowbreak Frequency\}$. 

\textbf{Argument role prediction:} The argument role layer predicts the assignment of arguments to triggers using separate binary classifiers, $d$, for each labeled argument and one classifier for all span-only arguments, $d \in \{1,2,\ldots5\}$ (4 for labeled arguments and 1 for span-only arguments). Argument role scores for trigger $j$ and argument $k$ using argument role classifier $d$ are calculated as 
\begin{equation}
\bm{\psi}_{d}(s_j,s_k) = \bm{w}_{r,d}\mbox{FFNN}_{r,d}([\bm{g}_j,\bm{g}_k])
\end{equation}
where $\bm{\psi}_{d}(s_j,s_k)$ is a vector of size $2$, $\mbox{FFNN}_{r,d}$ is a non-linear projection from size $4v_h$ to $v_r$, and $\bm{w}_{r,d}$ has size $2 \times v_r$.

\textbf{Span pruning:} To limit the time and space complexity of the pairwise argument role predictions, only the top-$K$ spans for each span classifier, $c$, are considered during argument role prediction. The span score is calculated as the maximum label score in $\phi_{c}$, excluding the $null$ label score.

\subsection{Model Configuration}
The model configuration was selected using 3-fold cross validation (CV) on the training set. Table \ref{event_hyperparameters} in the Appendix summarizes the selected configuration. Training loss was calculated by summing the cross entropy across all span and argument role classifiers. Models were implemented using the Python PyTorch module \citep{NEURIPS2019_9015}.

\subsection{Data Representation}
During initial experimentation, \textit{Symptom} \textit{Assertion} extraction performance was high for the \textit{absent} subtype and lower for \textit{present}. The higher \textit{absent} performance is primarily associated with the consistent presence of negation cues, like ``denies'' or ``no.'' While there are affirming cues, like ``reports'' or ``has,'' the \textit{present} subtype is often implied by a lack of negation cues. For example, an entire sentence could be ``Short of breath.'' To provide the \textit{Symptom} \textit{Assertion} span classifier with a more consistent span representation, we replaced each \textit{Symptom} \textit{Assertion} span (token indices) with the \textit{Symptom} trigger span in each event and found that performance improved. We extended this trigger span substitution approach to all labeled arguments (\textit{Assertion}, \textit{Change}, \textit{Severity}, and \textit{Test Status}) and found performance improved. By substituting the trigger spans for the labeled argument spans, trigger and labeled argument prediction is roughly treated as a multi-label classification problem, although the model does not constrain trigger and labeled argument predictions to be associated with the same spans. As previously discussed, the scoring routine does not consider the span indices of labeled arguments.

\subsection{Results}

\begin{table*}[htb!]
    \small
    \centering

\input{tables/extraction_performance.tex}
    \caption{Extraction performance}
    \label{extraction_performance}
\end{table*}

Table \ref{extraction_performance} presents the event extraction performance on the training set using CV and the withheld test set. Extraction performance is similar on the train and test sets, even though the training set has higher rates of COVID-19 positive notes. \textit{COVID} trigger extraction performance is very high (0.97 F1) and comparable to the annotator agreement (0.95 F1). The \textit{COVID} \textit{Assertion} performance (0.73 F1) is higher than \textit{Test Status} performance (0.62 F1), which is likely due to the more consistent \textit{Assertion} annotation. \textit{Symptom} trigger and \textit{Assertion} extraction performance is high (0.83 F1 and 0.79 F1, respectively), approaching the annotator agreement (0.86 F1 and 0.83 F1, respectively). \textit{Anatomy} extraction performance (0.61 F1) is lower than expected, given the high annotator agreement (0.81 F1). \textit{Duration} extraction performance is comparable to annotator agreement, and \textit{Frequency} extraction performance is lower than annotation agreement. \textit{Change}, \textit{Severity}, and \textit{Characteristics} extraction performance is low, again likely related to low annotator agreement for these cases.

Existing symptom extraction systems do not extract all of the phenomena in the CACT annotation scheme; however, we compared the performance of our Span-based Event Extractor to MetaMapLite++ for symptom identification and assertion prediction. MetaMapLite++ is an analysis pipeline that includes a UMLS concept extractor with assertion prediction \citep{yetisgen2016new}. Table \ref{umls_performance} presents the performance of MetaMapLite++ on the CACT test set. The spans associated with medical concepts in MetaMapLite++ differ slightly from our annotation scheme. For example, ``dry cough'' was extracted by MetaMapLite++ as a symptom, whereas our annotation scheme labels ``cough'' as the symptom and ``dry'' as a characteristic. To account for this difference, Table \ref{umls_performance} presents the performance of MetaMapLite++ for two trigger equivalence criteria: 1) \textit{exact match} for triggers is required, as defined in Section 3.3 and 2) \textit{any overlap} for triggers is considered equivalent.  The Span-based Event Extractor outperforms MetaMapLite++ for symptom identification precision (0.81 vs. 0.66), recall (0.85 vs. 0.67), and F1 (0.83 vs. 0.66), even when MetaMapLite++ is evaluated with the more relaxed \textit{any overlap} trigger scoring. The lower recall of MetaMapLite++ is partially the result of the UMLS not including symptom acronyms and abbreviations that frequently occur in our data, for example ``N/V/D'' for ``nausea, vomiting, and diarrhea.'' Table \ref{umls_performance} only reports the performance for the UMLS ``Sign or Symptom'' semantic type. When all UMLS semantic types are used, the recall improves; however, the precision is extremely low ($P = 0.02-0.03$) \footnote{In the UMLS, 15\% of the unique gold symptoms in the CACT training set are covered when only the ``Sign or Symptom'' semantic type is used. The UMLS coverage increases to 48\% when all semantic types are used, and the unique gold symptoms occur in 76 different UMLS semantic types.}.

\begin{table*}[htb!]
    \small
    \centering

\input{tables/umls_extraction_performance}

    \caption{MetaMapLite++ extraction performance for \textit{Symptom} \textit{trigger} and \textit{Assertion}}
    \label{umls_performance}
\end{table*}

Table \ref{assertion_performance} presents the assertion prediction performance for both systems, only considering the subset of predictions with exact trigger matches (i.e. assertion prediction performance is assessed without incurring penalty for trigger identification errors). The number of gold assertion labels (``\# Gold'') is greater for the Span-based Event Extractor, because more of the symptom triggers predictions are correct. The Span-based Event Extractor outperforms MetaMapLite++ in assertion prediction precision (0.95 vs. 0.81), recall (0.94 vs. 0.81), and F1 (0.94 vs. 0.81). MetaMapLite++'s lower performance is partly the result of differences between the distribution of assertion labels in CACT and the dataset used to train MetaMapLite++'s assertion classifier (2010 i2b2).

\begin{table*}[htb!]
    \small
    \centering

\input{tables/assertion_comparison}
    \caption{\textit{Symptom} \textit{Assertion} comparison for events with equivalent triggers (exact span match)}
    \label{assertion_performance}
\end{table*}

\ifsubfile
\bibliography{mybib}
\fi

%% file: tables/extraction_performance.tex
\newcolumntype{A}{ >{\raggedleft\arraybackslash} m{0.38in} }
\newcolumntype{B}{ >{\raggedleft\arraybackslash} m{0.65in} }
\newcolumntype{R}{ >{\raggedleft\arraybackslash} m{0.18in} }

\begin{tabular}{m{0.72in} m{0.71in} A R R R B R R R}
\toprule
\multirow{2}{*}{\textbf{Event type}} & \multirow{2}{*}{\textbf{Argument}} & \multicolumn{4}{c}{\textbf{Train-CV}}                   & \multicolumn{4}{c}{\hspace{0.27in}\textbf{Test}}                       \\ \cmidrule{3-10} 
                                     &                                    & \textbf{\# Gold} & \textbf{P} & \textbf{R} & \textbf{F1} & \textbf{\# Gold} & \textbf{P} & \textbf{R} & \textbf{F1} \\ \toprule
\multirow{3}{*}{COVID}               & Trigger                            &   3,931 & 0.95 & 0.97 & 0.96 & 1,497 & 0.96 & 0.97 & 0.97 \\        \cmidrule{2-10}                                                   
                                     & Assertion                          &   2,936 & 0.70 & 0.74 & 0.72 & 1,075 & 0.72 & 0.74 & 0.73 \\        
                                     & Test Status                        &   1,068 & 0.60 & 0.62 & 0.61 &   457 & 0.63 & 0.60 & 0.62 \\       \midrule                                                         
\multirow{8}{*}{Symptom}             & Trigger                            &  13,823 & 0.82 & 0.85 & 0.83 & 5,789 & 0.81 & 0.85 & 0.83 \\          \cmidrule{2-10}                                                 
                                     & Assertion                          &  13,833 & 0.77 & 0.79 & 0.78 & 5,791 & 0.77 & 0.80 & 0.79 \\        
                                     & Change                             &     739 & 0.45 & 0.03 & 0.06 &   341 & 0.45 & 0.05 & 0.09 \\           
                                     & Severity                           &     743 & 0.47 & 0.30 & 0.37 &   327 & 0.45 & 0.31 & 0.37 \\         \cmidrule{2-10}                                                   
                                     & Anatomy                            &   3,839 & 0.76 & 0.59 & 0.66 & 1,959 & 0.78 & 0.50 & 0.61 \\          
                                     & Characteristics                    &   3,145 & 0.59 & 0.26 & 0.36 & 1,441 & 0.66 & 0.25 & 0.36 \\        
                                     & Duration                           &   3,744 & 0.62 & 0.44 & 0.51 & 1,344 & 0.54 & 0.56 & 0.55 \\         
                                     & Frequency                          &     801 & 0.64 & 0.39 & 0.48 &   250 & 0.60 & 0.51 & 0.55 \\        \bottomrule                                    
\end{tabular}

%% file: tables/umls_extraction_performance.tex
\begin{tabular}{lllll}
\toprule
\textbf{Case}                               & \textbf{Agument} & \textbf{P}    & \textbf{R}    & \textbf{F1}   \\ \toprule

\multirow{2}{*}{Exact trigger match}        & Trigger         & 0.53           & 0.54          & 0.54                 \\  
                                            & Assertion       & 0.43           & 0.44          & 0.44                 \\ \midrule
\multirow{2}{*}{Any triggers overlap}       & Trigger         & 0.66           & 0.67          & 0.66                 \\
                                            & Assertion       & 0.54           & 0.55          & 0.54                 \\

\bottomrule
\end{tabular}

%% file: tables/assertion_comparison.tex
\begin{tabular}{lllll}
\toprule
\textbf{Model}                  & \textbf{\# Gold} & \textbf{P}    & \textbf{R}    & \textbf{F1}   \\ \toprule
MetaMap++     & 3,152           & 0.81 & 0.81 & 0.81 \\
Span-based Event Extractor      & 4,952           & 0.95 & 0.94 & 0.94 \\ \bottomrule
\end{tabular}

%% file: sections/covid_prediction.tex
\label{covid_prediction}


The creation of the CACT Corpus and the Span-based Event Extractor is motivated by our larger effort to explore the clinical presentation of diseases through the comprehensive representation of symptoms across multiple dimensions. This section utilizes a subset of the extracted information to predict positive COVID-19 infection status among individuals presenting to clinical settings for COVID-19 testing. This experimentation uses COVID-19 test results to distinguish between COVID-19 negative and COVID-19 positive patients with the goals of identifying the clinical presentation of COVID-19 and investigating the predictive power of symptoms. Improved understanding of the clinical presentation of COVID-19 has the potential to improve risk stratification of patients presenting for COVID-19 testing (by increasing or decreasing their pre-test probability), and thus guide diagnostic testing and clinical decision making.

\subsection{Data}
An existing clinical data set from the UW from January 2020 through May 2020 was used to explore the prediction of COVID-19 test results and identify the most prominent predictors of COVID-19. The data set represents 230K patients, including 28K patients with at least one COVID-19 PCR test result. The data set includes telephone encounters, outpatient progress notes, and emergency department (ED) notes, as well as structured data (demographics, vitals, laboratory results, etc.). 

For each patient in this data set, all of the COVID-19 tests with either a \textit{positive} or \textit{negative} result \underline{and} at least one note within the seven days preceding the test result were identified. Only COVID-19 tests with a note within the previous seven days are included in experimentation, to improve the robustness of the COVID-19 symptomology exploration. Each of these test results was treated as a sample in this binary classification task (positive or negative). The likelihood of COVID-19 positivity was predicted using structured EHR data and notes within a 7-day window preceding the test result. The pairing of notes and COVID-19 test results was independently performed for each of the note types (ED, outpatient progress, and telephone encounter notes). From this pool of data, we identified the following test counts by note type: 2,226 negative and 148 positive for ED; 7,599 negative and 381 positive for progress; and 7,374 negative and 448 positive for telephone. Within the 7-day window of this subset of COVID-19 test results, there are  5.3K ED, 14.5K progress, and 27.5K telephone notes. This data set has some overlap with the data set used in Section \ref{methods_section} but is treated as a separate data set in this COVID-19 prediction task. The notes in the CACT training set are less than 1\% of the notes used in this secondary use application.

\subsection{Methods}

\textbf{Features:} Symptom information was automatically extracted from the notes using the Span-based Event Extractor trained on CACT.\footnote{Only automatically extracted symptom data were used. No supervised (hand annotated) labels were used.} The extracted symptoms were normalized using the mapping in Table \ref{symptom_normalization} in the Appendix. Each extracted symptom with an \textit{Assertion} value of ``present'' was assigned a feature value of 1. The 24 identified predictors of COVID-19 from existing literature (see Section \ref{lit_review}) were mapped to 32 distinct fields within the UW EHR and used in experimentation. Identified fields are listed in Table \ref{ehr_features} of the Appendix. For the coded data (e.g. structured fields like ``basophils''), experimentation was limited to this subset of literature-supported COVID-19 predictors, given the limited number of positive COVID-19 tests in this data set. 

Within the 7-day history, features may occur multiple times (e.g. multiple temperature measurements). For each feature, the series of values was represented as the minimum or maximum of the values depending on the specific feature. For example, temperature was represented as the maximum of the measurements to detect any fever, and oxygen saturation was represented as the minimum of the values to capture any low oxygenation events. Table \ref{ehr_features} in the Appendix includes the aggregating function, $\bm{f}$, used for each field.

Where symptom features were missing, the feature value was set to 0. For features from the structured EHR data, which are predominantly numerical, missing features were assigned the mean feature value in the set used to train the COVID-19 prediction model. 

\textbf{Model:} COVID-19 was predicted using the Random Forest framework, because it facilitates nonlinear modeling with interdependent features and interpretability analyses (Scikit-learn Python implementation used \citep{scikit-learn}). Alternative prediction algorithms include Logistic Regression, SVM, and FFNN. Logistic Regression assumes feature independence and linearity, which is not valid for this task. For example, the feature set includes both the symptom ``fever'' and temperature measurements (e.g. ``$38.6^{\circ}C$''). Model interpretability is less clear with SVM, and the number of positive test samples is relatively small for a FFNN.

The relative importance of features in predicting COVID-19 was explored using \citet{lundberg2020local}’s SHAP (SHapley Additive exPlanations) approach, which is implemented in the SHAP Python module.\footnote{\url{https://pypi.org/project/shap/}} SHAP generates interpretable, feature-level explanations for nonlinear model predictions. For each prediction, SHAP feature scores are estimated, where larger absolute scores indicate higher importance, and the absolute values of the scores sum to 1.0 for each prediction. 

\textbf{Experimental paradigm:} The available data was split into train/test sets using an 80\%/20\% split by patient, although training and evaluation was performed at the test-level (i.e. each COVID-19 test result is a sample). Performance was evaluated using the receiver operating characteristic (ROC) and the associated area under the curve (AUC). Given the relatively small number of positive samples, the train/test splits were randomly created 1,000 times through repeated hold-out testing \citep{kim2009estimating}. \citet{kim2009estimating} demonstrated that repeated hold-out testing can improve the robustness of the results in low resource settings. For each train/test split, the AUC was calculated, and an average AUC was calculated across all hold-out iterations. The random holdout iterations yield a distribution of AUC values, which facilitate significance testing. The significance of the AUC performance was assessed using a two-sided T-test. The Random Forest models were tuned using 3-fold cross validation on the training set and evaluated on the withheld test set. COVID-19 prediction experimentation included three feature sets: \textit{structured} (32 structured EHR fields), \textit{notes} (automatically extracted symptoms), and \textit{all} (combination of structured fields and automatically extracted symptoms). Separate models were trained and evaluated for each note type (ED, progress, and telephone) and feature set (\textit{structured}, \textit{notes}, and \textit{all}). The selected Random Forest hyperparameters are summarized in Table \ref{covid_hyperparams} in the Appendix.

\subsection{Results}
Figure \ref{roc} presents the ROC for the COVID-19 predictors with the average AUC across repeated hold-out partitions. The AUC evaluates model performance across all operating points, including operating points that are not clinically significant, for example extremely low true positive rate (TPR). To address this AUC limitation and provide an alternative method for comparing feature sets, we selected a fixed operating point on the ROC, comparing the false positive rate (FPR) at a specific TPR. We selected a TPR (sensitivity) of 80\%, as a value that has clinical value for identifying individuals with COVID-19, and we examined the FPR (specificity) at this fixed TPR. In this use case, we are attempting to see how well structured EHR fields and symptoms perform compared to the reference standard of a laboratory PCR test. 

Table \ref{fpr_at_tpr} presents the FPR at TPR=0.80, including the FPR mean and standard deviation across the repeated holdout iterations. Lower FPRs (better performance) are achieved for all three note types, when automatically extracted symptoms are added to the structured data. We would not expect a combination of clinical features to have particularly high sensitivity. \citet{smith2020symptom} achieved similar performance in predicting COVID-19 using clinical prediction rules. While detecting COVID-19 in 80\% of patients with the disease, the inclusion of automatically extracted symptoms decreases the FPR (the ``cost'') by 2-7 percentage points. For all note types, the inclusion of the automatically extracted symptom information (\textit{all} feature set) improves performance  over structured data only (\textit{structured}-only feature set) for both AUC and FPR$@$TPR=0.80 with significance ($p<0.001$ per two-sided T-test). The \textit{structured} features achieve higher performance in the ED note experimentation, than experimentation with progress and telephone notes, due to the higher prevalence of vital sign measurements and laboratory testing in proximity to ED visits. In ED note experimentation, over 99\% of samples include vital signs and 72\% include blood work. In progress and telephone note experimentation, 23-38\% of samples includes vital signs and 19-26\% include blood work. The automatically extracted symptoms are especially important in clinical contexts, like outpatient and tele-visit, where vital signs, laboratory results, and other structured data are less available.

\begin{figure}[t!]
    \centering
    \iftif
        \includegraphics[width=6.0in]{figures/roc.tif}
    \else
        \includegraphics[width=6.0in]{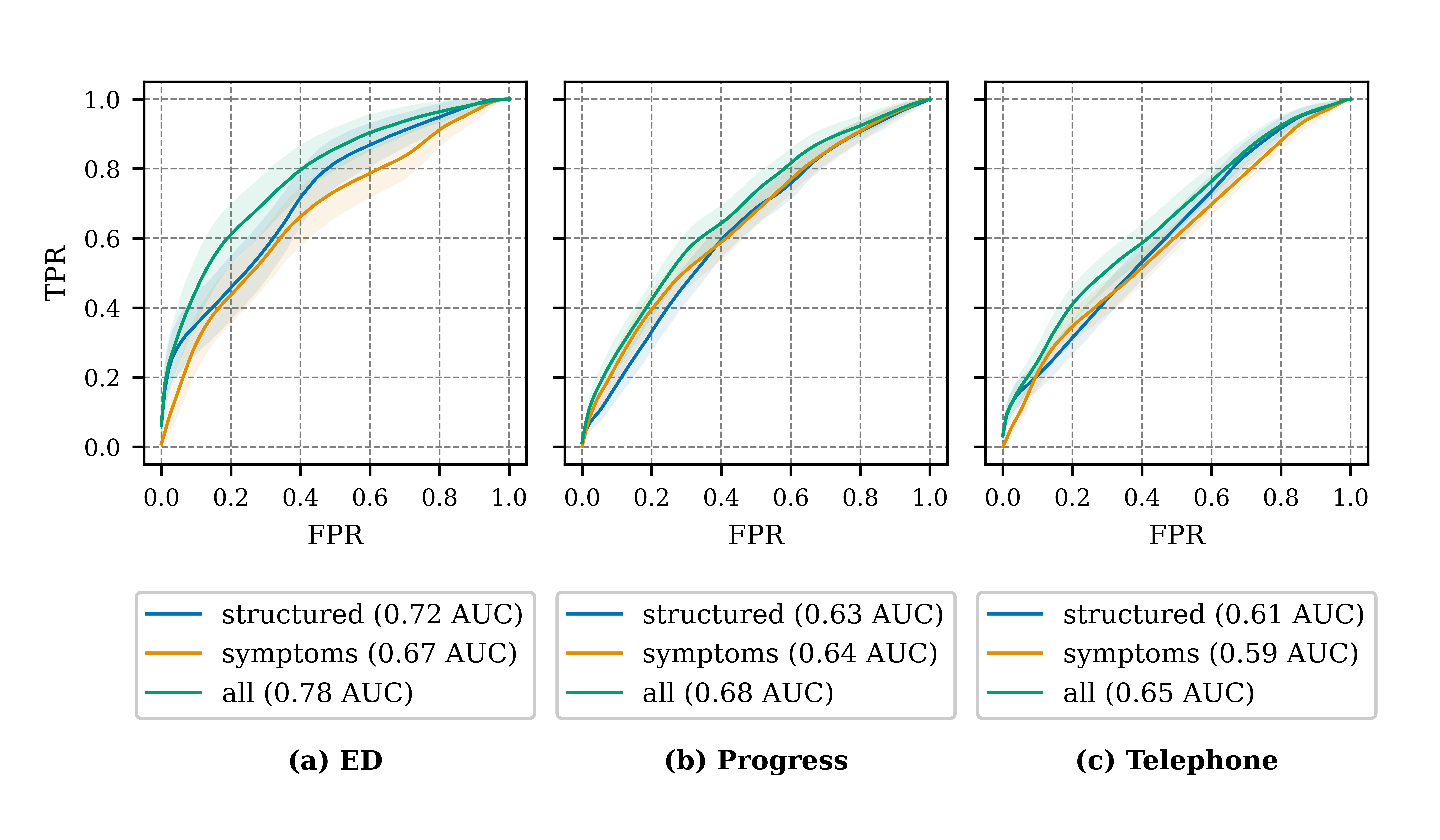}   
    \fi
    \caption{Receiver operating characteristic by note type and feature set combination for repeated hold-out iterations. The solid line indicates the average ROC, and the shaded region around the solid line indicates one standard deviation.}
    \label{roc}
\end{figure}

\begin{table}[h]
    \small
    \centering

\input{tables/fpr_at_tpr}
    \caption{COVID-19 prediction false positive rate at a true positive rate of 80\%}
    \label{fpr_at_tpr}    
\end{table}

\begin{figure}[ht!]
    \centering
    \iftif
        \includegraphics[width=5.0in]{figures/shap_impact.tif}
    \else
        \includegraphics[width=5.0in]{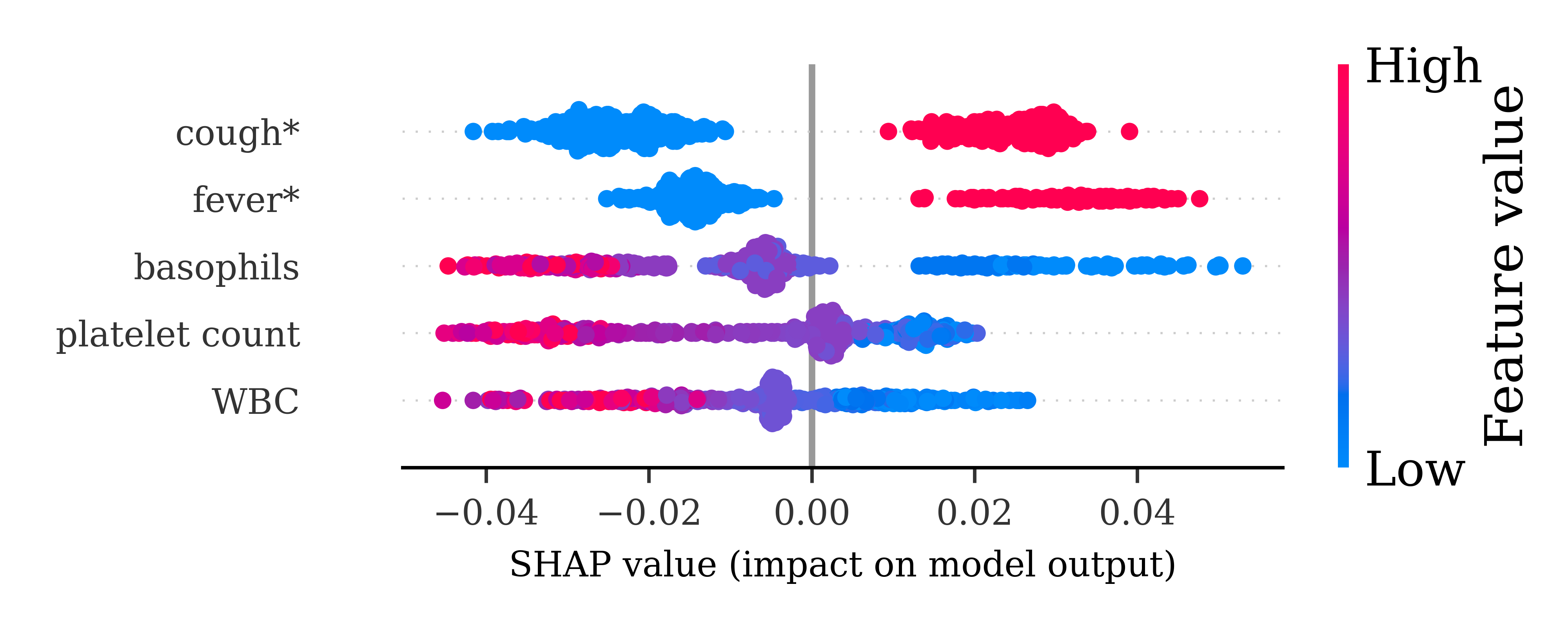}       
    \fi
    \caption{SHAP plot for a single Random Forest model from the ED note experimentation with the \textit{all} feature set, explaining the importance of features in making predictions for the withheld test set. * indicates the feature is an automatically extracted symptom}
    \label{shap_impact_single}
\end{figure}

Figure \ref{shap_impact_single} presents a SHAP value plot for the five most predictive features from a single Random Forest model from the ED note experimentation with the \textit{all} feature set. In this SHAP plot, each point represents a single test prediction, and the SHAP value (x-axis) describes the feature importance. Positive SHAP values indicate support for COVID-19 positivity, and negative values indicate support for negative test result. The color coding indicates the feature value, where red indicates higher feature values and blue indicates lower feature values. For example, high and moderate basophils values (coded in red and purple, respectively) have negative SHAP values, indicating support COVID-19 negativity. Low basophils values (coded in blue) have positive SHAP values, indicating support COVID-19 positivity. 

\begin{figure}[t!]
    \centering

    \iftif
        \includegraphics[width=6.0in]{figures/shap_violin.tif}
    \else
        \includegraphics[width=6.0in]{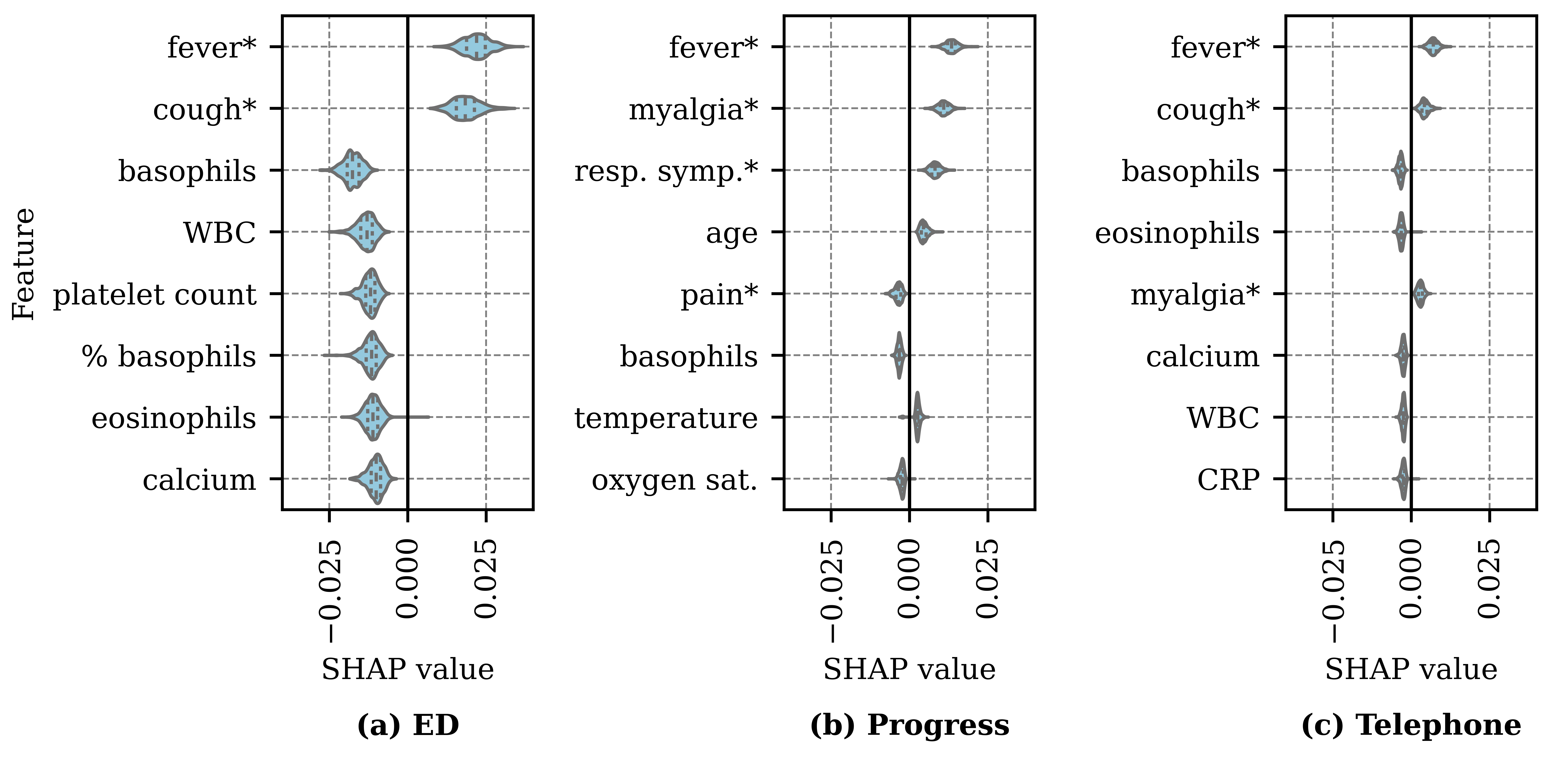}   
    \fi    
    
    \caption{Distribution of averaged SHAP values by note type with the \textit{all} feature set. The vertical lines in each violin indicate the quartiles. * indicates the feature is an automatically extracted symptom.}
    \label{shap_impact_violin}
\end{figure}

Given the relatively small sample size and low proportion of positive COVID-19 tests, the SHAP impact values presented in Figure \ref{shap_impact_single} were aggregated across repeated hold-out runs. Figure \ref{shap_impact_violin} presents the averaged SHAP values for each repeated hold-out run for the eight most predictive features for the \textit{all} feature set. For each repeated hold-out run, the absolute value of the SHAP values were averaged, yielding a single feature score per repetition. The mean SHAP values (x-axis) represents the importance of the feature in predicting COVID-19, where positive values indicate a positive correlation between the feature values and COVID-19 positivity and negative values indicate a negative correlation. The most predictive features vary by note type, although fever is a prominent indicator of COVID-19 across note types. For each note type, the top five symptoms indicating COVID-19 positivity include: ED - fever, cough, myalgia, fatigue, and flu-like symptoms; progress - fever, myalgia, respiratory symptoms, cough, and ill; and telephone - fever, cough, myalgia, fatigue, and sore throat. The differences in symptom importance by note type reflects differences in documentation in the clinical settings (e.g., emergency department, outpatient, and tele-visit).

\ifsubfile
\bibliography{mybib}
\fi

%% file: tables/fpr_at_tpr.tex
\begin{tabular}{lll}
\toprule
\textbf{Note type} & \textbf{Feature type} & \textbf{FPR @ TPR = 0.80}       \\ \midrule
\multirow{3}{*}{ED}        & structured & 0.48 $\pm$ 0.09 \\
                           & symptoms   & 0.63 $\pm$ 0.12 \\
                           & all        & 0.41 $\pm$ 0.10  \\ \midrule
\multirow{3}{*}{Progress}  & structured & 0.64 $\pm$ 0.05 \\
                           & symptoms   & 0.64 $\pm$ 0.05 \\
                           & all        & 0.58 $\pm$ 0.06 \\   \midrule
\multirow{3}{*}{Telephone} & structured & 0.66 $\pm$ 0.04 \\  
                           & symptoms   & 0.71 $\pm$ 0.03 \\
                           & all        & 0.64 $\pm$ 0.05 \\ \bottomrule                                    
\end{tabular}

%% file: sections/conclusions.tex
We present CACT, a novel corpus with detailed annotations for COVID-19 diagnoses, testing, and symptoms. CACT includes 1,472 unique notes across six note types with more than 500 notes from patients with future positive COVID-19 tests. We implement the Span-based Event Extractor, which jointly extracts all annotated phenomena, including argument types and subtypes. The Span-based Event Extractor achieves near-human performance in the extraction of \textit{COVID} triggers (0.97 F1) and \textit{Symptom} triggers (0.83 F1) and \textit{Assertions} (0.79 F1). The performance of several attributes (e.g. \textit{Change}, \textit{Severity}, \textit{Characteristics}, \textit{Duration}, and \textit{Frequency}) is lower than that of \textit{Assertion}. This lower performance may partly be due to the focus on COVID-19, where clinicians’ notes: 1) are highly structured around the presence/absence of a certain set of symptoms, 2) usually describe a single consultation per patient within 7 days of COVID-19 testing, and 3) focus on assessing the need for COVID-19 testing and in-person ambulatory or ED care.  

In a COVID-19 prediction task, automatically extracted symptom information improved the prediction of COVID-19 test results (with significance) beyond just using structured data, and the top predictive symptoms include fever, cough, and myalgia. This application is limited by the size and scope of the available data. CACT only includes notes from early in the COVID-19 pandemic (February-March 2020), and our understanding of the presentation of COVID-19 has evolved since that time. CACT was annotated for all symptoms described in the clinical narrative, not just known symptoms of COVID-19, so the annotated symptoms cover most of the symptoms currently known to be associated with COVID-19. However, CACT includes infrequent references to losses of taste or smell. Additional annotation of notes from later in the pandemic is needed to address this gap.

In future work, the extractor will be applied to a much larger set of clinical ambulatory care and ED notes from UW. The extracted symptom information will be combined with routinely coded data (e.g. diagnosis and procedure codes, demographics) and automatically extracted data (e.g. social determinants of health \citep{lybarger2020annotating}). Using these data, we will develop models for predicting risk of COVID-19 infection. These models could better inform clinical indications for prioritizing testing, and the presence or absence of certain symptoms can be used to inform clinical care decisions with greater precision. This future work may also identify combinations of symptoms (including their presence, absence, severity, sequence of appearance, duration, etc.) associated with clinical outcomes and health service utilization, such as deteriorating clinical course and need for repeat consultation or hospital admission. The use of detailed symptom information will be highly valuable in informing these models, but potentially only with the level of nuance that our extraction models provide. For the COVID-19 pandemic, we anticipate that the extraction model presented here will be of increasing value to clinical researchers, as the need to distinguish COVID-19 from other viral and bacterial respiratory infections becomes more necessary. As the pandemic subsides with widespread vaccination, we will return to the more typical ``winter respiratory infection/influenza'' seasons, where routine medical care involves differentiating COVID-19 from many other types of viral infections and identifying individuals that require COVID-19 testing. Symptom extraction models, like the model presented here, may provide the data needed to determine risk of certain infections and triage the need for testing. 

We intend to explore the value of this symptom annotation scheme and extraction approach for clinical conditions where multiple consultations lead to a time point in the diagnosis pathway and symptom attributes, like change and severity, are even more important. We are especially interested in medical conditions for which delayed or missed diagnoses are known to lead to patient harm \cite{newman2019serious}. We intend to examine data sets associated with other acute and chronic conditions to investigate symptom patterns that could be used to more efficiently and accurately identify patients with these conditions. Specifically, we intend to further develop the symptom extractor to reduce diagnostic delay for lung cancer, which is known to present at a later stage. Lung cancer diagnosis often occurs after many consultations in ambulatory settings, and there may be opportunities to more quickly identify high-risk individuals based on symptoms.

\ifsubfile
\bibliography{mybib}
\fi

%% file: sections/appendix.tex
\input{tables/symptom_normalization.tex}

\begin{table}[ht!]
\small
\centering
\caption{Demographic, vital signs, and laboratory fields that are predictive of COVID-19 infection in current literature}
\label{lit_features}

\input{tables/lit_review_features.tex}

\end{table}

\begin{table}[htb]
    \small
    \centering
    \caption{Hyperparameters for the Span-based Event Extractor}

\input{tables/hyperparams}
    \label{event_hyperparameters}
\end{table}

\begin{table}[ht!]
\small
\centering
\caption{Structured fields from UW EHR used to predict COVID-19 infection. $\bm{f}$ indicates the function used to aggregate multiple measurements/values. Fields that measure the same phenomena and were treated as a single feature, resulting in 29 distinct structured EHR fields: \{``Temperature - C,'' ``Temperature (C)''\}, \{``HR,'' ``Heart Rate''\}, and \{``O2 Saturation (\%),'' ``Oxygen Saturation''\}. All fields numerical (e.g. “Temperature (C)”=38.1), except “Troponin I Interpretation” and "Gender"}
\label{ehr_features}

\input{tables/uw_ehr_features.tex}

\end{table}

\begin{table}[ht!]
\small
\centering
\caption{COVID-19 prediction hyperparameters for Random Forest Models}
\label{covid_hyperparams}

\input{tables/covid_hyperparams.tex}
\end{table}

\ifsubfile
\bibliography{mybib}
\fi

%% file: tables/symptom_normalization.tex
{\small

\renewcommand*{\arraystretch}{0.85}
\begin{longtable}{ p{1.2in} p{4.2in}}
\caption{Expert-derived mapping of symptoms to canonical forms} \\
\label{symptom_normalization} \\
\toprule
\multicolumn{1}{l}{\textbf{Normalized symptom}}   & \multicolumn{1}{l}{\textbf{Symptom variants}} \\ \toprule \endfirsthead

\multicolumn{2}{l}{Table \ref{symptom_normalization} - continued from previous page}\\
\hline \multicolumn{1}{l}{\textbf{Normalized symptom}} & \multicolumn{1}{l}{\textbf{Symptom variants}} \\ \toprule
\endhead

\hline \multicolumn{2}{r}{{Continued on next page}} \\ \bottomrule
\endfoot

\bottomrule
\endlastfoot

altered mental status	&	ams, confused, confusion	\\ \midrule
anxiety	&	agitated, agitation, anxious	\\ \midrule
arthralgia	&	arthralgias	\\ \midrule
bleeding	&	bleed, blood, bloody	\\ \midrule
bruising	&	bruise, bruises, ecchymosis	\\ \midrule
chest pain	&	cp	\\ \midrule
chills	&	chill	\\ \midrule
cough	&	c, c., cough cough, coughing, coughs, distress coughing, distressed coughing	\\ \midrule
cramping	&	cramps	\\ \midrule
decreased appetite	&	loss of appetite, poor appetite, poor p.o. intake, poor po intake, reduced appetite	\\ \midrule
deformities	&	deformity	\\ \midrule
dehydration	&	dehydrated	\\ \midrule
diarrhea	&	d, d., diarrhea stools, loose stools	\\ \midrule
disharge	&	drainage	\\ \midrule
distended	&	distention	\\ \midrule
dysphagia	&	difficulty swallowing, dysphagia symptoms	\\ \midrule
erythema	&	erythematous, redness	\\ \midrule
exudates	&	exudate	\\ \midrule
fall	&	falls	\\ \midrule
fatigue	&	drowsiness, drowsy, fatigued, somnolence, somnolent, tired, tiredness	\\ \midrule
fever	&	f, f., febrile, fevers	\\ \midrule
flu-like symptoms	&	flu - like symptoms, influenza - like symptoms	\\ \midrule
gi symptoms	&	abdominal symptoms	\\ \midrule
headache	&	ha, headaches	\\ \midrule
heartburn	&	gerd symptoms, heartburn symptoms	\\ \midrule
hematochezia	&	brbpr	\\ \midrule
ill	&	ill - appearing, ill appearing, ill symptoms, illness, sick	\\ \midrule
incontinent	&	incontinence	\\ \midrule
irritation	&	irritable	\\ \midrule
itching	&	itchy	\\ \midrule
lethargy	&	lethargic	\\ \midrule
lightheadedness	&	dizziness, dizzy, headedness, lightheaded	\\ \midrule
myalgia	&	ache, aches, aching, bodyaches, myalgias	\\ \midrule
nausea	&	n, n., nauseated, nauseous	\\ \midrule
pain	&	discomfort, painful, pains	\\ \midrule
pruritus	&	pruritis	\\ \midrule
rash	&	rashes	\\ \midrule
respiratory symptoms	&	uri symptoms	\\ \midrule
runny nose	&	rhinorrhea	\\ \midrule
seizures	&	seizure, seizures	\\ \midrule
shortness of breath	&	\_\_\_shortness of breath, difficult breathing, difficulty breathing, difficulty of breathing, distress breathing, distressed breathing, doe, dsypnea, dypsnea, dyspnea, dyspnea exertion, dyspnea on exertion, increase work of breathing, increased work of breathing, out of breath, respiratory distress, short of breath, shortneses of breath, shortness breath, shortness of breaths, sob, sob on exertion, trouble breathing, work of breathing	\\ \midrule
sore throat	&	pharyngitis	\\ \midrule
soreness	&	sore	\\ \midrule
sputum	&	sputum production	\\ \midrule
sweats	&	diaphoresis, nightsweats, sweating	\\ \midrule
swelling	&	edema, oedema, swollen	\\ \midrule
syncope	&	fainting	\\ \midrule
tenderness	&	tender	\\ \midrule
tremors	&	tremor	\\ \midrule
ulcers	&	ulcer, ulceration, ulcerations	\\ \midrule
urinary symptoms	&	urinary	\\ \midrule
urination	&	urinating	\\ \midrule
vomiting	&	emesis, v, v., vomitting	\\ \midrule
weakness	&	weak	\\ \midrule
wheezing	&	wheeze, wheezes	\\ \midrule
wounds	&	wound	\\ \bottomrule

\end{longtable}
}



%% file: tables/lit_review_features.tex
\begin{tabular}{ll}
\toprule
\textbf{Parameter}                  & \textbf{Sources}                                                       \\ \midrule
age                                 & \citep{bertsimas2020predictions, Wynantsm1328, Brinati2020, Mei2020} \\ 
alanine aminotransferase (ALT)      & \citep{SIORDIA2020104357, Brinati2020} \\ 
albumin                             & \citep{zhang2020risk} \\ 
alkaline phosphatase (ALP)          & \citep{Brinati2020} \\ 
aspartate aminotransferase (AST)    & \citep{bertsimas2020predictions, SIORDIA2020104357, Brinati2020} \\ 
basophils                           & \citep{Brinati2020} \\ 
calcium                             & \citep{bertsimas2020predictions} \\ 
C-reactive protein (CRP)            & \citep{bertsimas2020predictions, SIORDIA2020104357, zhang2020risk} \\ 
D-dimer                             & \citep{ SIORDIA2020104357} \\ 
eosinophils                         & \citep{SIORDIA2020104357, Brinati2020} \\ 
gamma-glutamyl transferase (GGT)    & \citep{Brinati2020} \\ 
gender                              & \citep{Mei2020} \\ 
heart rate                          & \citep{bertsimas2020predictions} \\ 
lactate dehydrogenase (LDH)         & \citep{ SIORDIA2020104357, zhang2020risk, Brinati2020} \\ 
lymphocytes                         & \citep{Wynantsm1328, SIORDIA2020104357, zhang2020risk, Brinati2020, Mei2020} \\ 
monocytes                           & \citep{Brinati2020} \\ 
neutrophils                         & \citep{ Wynantsm1328, SIORDIA2020104357, Brinati2020, Mei2020} \\ 
oxygen saturation                   & \citep{bertsimas2020predictions} \\ 
platelets                           & \citep{Brinati2020} \\ 
prothrombin time (PT)               & \citep{ SIORDIA2020104357} \\ 
respiratory rate                    & \citep{bertsimas2020predictions} \\ 
temperature                         & \citep{bertsimas2020predictions, Wynantsm1328, Mei2020} \\ 
troponin                            & \citep{ SIORDIA2020104357} \\ 
white blood cell (WBC) count        & \citep{bertsimas2020predictions, Brinati2020, Mei2020} \\ 

\bottomrule
\end{tabular}

%% file: tables/hyperparams.tex
\begin{tabular}{m{1.7in} >{\raggedleft\arraybackslash} m{1.1in} }
\toprule
\textbf{Parameter}                  & \textbf{Value} \\ \midrule
Maximum sentence length, $n$        & 30             \\ 
Maximum span length, $M$            & 6              \\ 
Top-$K$ spans per classifier        & sentence token count            \\ 
Batch size                          & 100            \\ 
Number of epochs                    & 100            \\ 
Learning rate                       & 0.001          \\ 
Optimizer                           & Adam           \\ 
Maximum gradient L2-norm            & 100            \\ 
BERT embedding dropout              & 0.3            \\ 
bi-LSTM hidden size, $v_h$          & 200            \\ 
bi-LSTM activation function         & tanh           \\ 
bi-LSTM  dropout                    & 0.3            \\ 
Span classifier projection size, $v_s$ & 100         \\ 
Span classifier activation function & ReLU           \\ 
Span classifier dropout             & 0.3            \\ 
Role classifier projection size, $v_r$ & 100         \\ 
Role classifier activation function & ReLU           \\ 
Role classifier dropout             & 0.3            \\ \bottomrule
\end{tabular}

%% file: tables/uw_ehr_features.tex
\begin{tabular}{ p{1.2in} p{3.0in} p{0.3in} }
\toprule
\textbf{Parameter}  & \textbf{Fields in UW EHR}                             & $\bm{f}$  \\ \toprule
age                 & ``AgeIn2020''                                         & max       \\ \midrule
ALT                 & ``ALT (GPT)''                                         & max       \\ \midrule
albumin             & ``Albumin''                                           & min       \\ \midrule
ALP                 & ``Alkaline Phosphatase (Total)''                      & max       \\ \midrule
AST                 & ``AST (GOT)''                                         & max       \\ \midrule
basophils           & ``Basophils'' and ``\% Basophils''                    & min       \\ \midrule
calcium             & ``Calcium''                                           & min       \\ \midrule
CRP                 & ``CRP, high sensitivity''                             & max       \\ \midrule
D-dimer             & ``D\_Dimer Quant''                                    & max       \\ \midrule
eosinophils         & ``Eosinophils'' and ``\% Eosinophils''                & min       \\ \midrule
GGT                 & ``Gamma Glutamyl Transferase''                        & max       \\ \midrule
gender              & ``Gender''                                            & last      \\ \midrule
heart rate          & ``Heart Rate'' and ``HR''                             & max       \\ \midrule
LDH                 & ``Lactate Dehydrogenase''                             & max       \\ \midrule
lymphocytes         & ``Lymphocytes'' and ``\% Lymphocytes''                & min       \\ \midrule
monocyptes          & ``Monocytes''                                         & max       \\ \midrule
neutrophils         & ``Neutrophils'' and ``\% Neutrophils''                & max       \\ \midrule
oxygen saturation   & ``Oxygen Saturation'' and ``O2 Saturation (\%)''      & min       \\ \midrule
platelets           & ``Platelet Count''                                    & min       \\ \midrule
PT                  & ``Prothrombin Time Patient'' and ``Prothrombin INR''  & max       \\ \midrule
respiratory rate    & ``Respiratory Rate''                                  & max       \\ \midrule
temperature         & ``Temperature - C'' and ``Temperature (C)''           & max       \\ \midrule
troponin            & ``Troponin\_I'' and ``Troponin\_I Interpretation''    & max       \\ \midrule
WBC count           & ``WBC''                                               & min       \\ \bottomrule
\end{tabular}

%% file: tables/covid_hyperparams.tex
\begin{tabular}{m{0.65in} m{0.65in} m{0.65in} m{0.65in} m{0.65in} m{0.65in} m{0.65in}}
\toprule
\textbf{Note type}    & \textbf{Features}      &	\textbf{\# estimators}   & \textbf{Maximum depth}     & \textbf{Minimum samples per split}     & \textbf{Minimum samples per leaf}  &     	\textbf{Class weight ratio (pos./neg.)} \\ \midrule
ED           & 	structured   & 	200	            &    4	            &   2	                        &   1	                    &    10  \\
ED       	 &   notes    	 &  200	            &    4	            &   4	                        &   1	                    &    6   \\
ED   	     &   all  	     &  200	            &    6	            &   3	                        &   1	                    &    8   \\ \midrule
Progress     & 	structured   & 	200	            &    18	            &   6	                        &   1	                    &    6   \\
Progress 	 &   notes    	 &  200	            &    10	            &   4	                        &   1	                    &    4   \\
Progress     &  all          &  200	            &    8	            &   4	                        &   1	                    &    6   \\ \midrule
Telephone    & 	structured   & 	200	            &    10	            &   2	                        &   1	                    &    2   \\
Telephone	 &   notes    	 &  200	            &    6	            &   2	                        &   1	                    &    4   \\
Telephone    &	all          &	200	            &    10	            &   8	                        &   1	                    &    4   \\ \bottomrule
\end{tabular}
             